\documentclass[3p,times]{elsarticle}
\usepackage[utf8]{inputenc}
\usepackage{graphicx}
\usepackage{float}
\usepackage{multirow}
\usepackage{adjustbox}
\usepackage{subfigure}
\usepackage{xcolor}
\usepackage{appendix}
\newtheorem{definition}{Definition}
\usepackage{amssymb}

\usepackage[figuresright]{rotating}
\usepackage{framed}
\usepackage{nomencl}
\makenomenclature
\begin{document}

\begin{frontmatter}

\title{Physics-Infused Fuzzy Generative Adversarial Network for Robust Failure Prognosis}
\author{Ryan Nguyen}
\ead{rnguye2@clemson.edu}
\address{Department of Automotive Engineering\\
       Clemson University\\
       Greenville, SC 29616, USA}
\date{November 2021}
\author{Shubhendu Kumar Singh}
\ead{shubhes@clemson.edu}
\address{Department of Automotive Engineering\\
       Clemson University\\
       Greenville, SC 29616, USA}
\author{Rahul Rai\corref{Rahul Rai}}
\cortext[Rahul Rai]{Corresponding author}
\ead{rrai@clemson.edu}
\address{Department of Automotive Engineering\\
       Clemson University\\
       Greenville, SC 29616, USA}

\begin{abstract}
Prognostics aid in the longevity of fielded systems or products. Quantifying the system's current health enable prognosis to enhance the operator's decision-making to preserve the system's health. Creating a prognosis for a system can be difficult due to (a) unknown physical relationships and/or (b) irregularities in data appearing well beyond the initiation of a problem. Traditionally, three different modeling paradigms have been used to develop a prognostics model: physics-based (PbM), data-driven (DDM), and hybrid modeling. Recently, the hybrid modeling approach that combines the strength of both PbM and DDM based approaches and alleviates their limitations is gaining traction in the prognostics domain. 
In this paper, a novel hybrid modeling approach for prognostics applications based on combining concepts from fuzzy logic and generative adversarial networks (GANs) is outlined.
The FuzzyGAN based method embeds a physics-based model in the aggregation of the fuzzy implications. This technique constrains the output of the learning method to a realistic solution.
Results on a bearing problem showcases the efficacy of adding a physics-based aggregation in a fuzzy logic model to improve GAN's ability to model health and give a more accurate system prognosis.
\end{abstract}

\end{frontmatter}
\newpage

\section{Introduction}
Prognostics is the study of predicting when a system or a component will no longer perform its intended function. The time between the current measured state and the end-of-life (EOL) is considered the remaining useful life (RUL). Challenges in acquiring RUL from the current state vary depending on the approach taken to model the system. Three different system modeling approaches are used in the prognostics domain: physics-based (PbM), data-driven (DDM), and hybrid modeling. 

PbM is based on observing the system behavior or deriving interactions from first principles \cite{parlos2000multi}. An example of PbM for prognostics is presented by Qiu et al. \cite{qiu2002damage}--they outline a physics-based model using damage mechanics and incorporate a vibration signal analysis method to complete the model.  However, PbM poses three significant challenges: (1) the incompleteness of a system model used for prognostics due to abstraction and model simplification, (2) the inability to assimilate information provided by sensors to estimate the parameters of the system model accurately, and (3) the high computational cost of running simulations on a highly nonlinear multi-physics system having multiple operating modes. These challenges lead to shortcomings of PbM for RUL: failing to fully capture the complex characteristics of system behavior due to the modeling abstractions--resulting in low fidelity and reduced-order models, or becoming computationally prohibitive due to the high computational cost associated with high-fidelity simulations.  Various approaches to developing more complete models are reviewed by Cubillo et al. \cite{cubillo2016review}.

DDM based prognostics methods can be divided into statistical models, feature extraction methods, and deep learning (DL) methods. Statistical models assume that the degradation processes are affected by variability both individually and temporally and have shown promising results for predicting RUL with less data than DL models. Gao et al. \cite{gao2019wiener} use the Wiener process for RUL prediction of light-emitting diode driving power in rail vehicle carriages. Furthermore, Sun et al. \cite{sun2021improved} use an inverse Gaussian process for RUL prediction of hydraulic piston pump, and Dong et al. \cite{dong2019data} develop a statistical model for battery health prognosis using an adaptive Brownian motion model. Typically, these statistical models separate the parameter identification and drift estimation process, which might result in poor convergence, accuracy, and robustness that DL models do not suffer from, Dong et. al \cite{dong2019data} address this issue through their adaptive Brownian motion approach. However, they assume that a health index is known prior--when the actual battery health index is hard to measure or estimate. Their degradation model was also simplified to use only one quality degradation characteristic. These shortcomings provide an opening for models outside the statistical model domain to be explored. Feature extraction methods use the data provided about the system and attempt to extract features using signal processing and data conditioning techniques. Hite \cite{hite1993algorithm} outlines a list of useful features to extract when establishing the health of a bearing system--such as the ball pass frequency of the inner and outer race. Additionally, Wang et al. \cite{wang2016size}, and Sawalhi \cite{sawalhi2017vibration} focus on spall size for determining the RUL of a bearing system. The authors of both papers propose a scheme of estimating bearing spall size based on synchronous signal averaging (SSA) concerning the bearing fault characteristic frequency, combined with envelope and wavelet analyses of the averaged signals. Another feature extraction method uses trajectory-based similarity prediction (TBSP) \cite{soons2020predicting, huang2019improved}. TBSP consists of finding previous runs similar to the current run to be predicted. A $k$ number of runs is found, and the average total life of those runs is taken; then, the difference between the current and averaged values is the RUL. For RUL estimation, the parameters that govern the RUL usually belong to a PbM but cannot be determined without further data conditioning. Once the data is conditioned, determining the missing parameters is usually done through manual inspection of the data; however, this is highly inefficient. On the contrary, one can use an automated approach to inspect particular data behaviors, but those non-DL approaches are prone to error.

DL has demonstrated the ability to identify specific trends in data and excels in image processing, natural language understanding, and Big Data analytics \cite{pinheiro2015learning, kumar2016ask}. Recently, DL-based approaches have been used to predict the RUL of a system. Recent publications \cite{lu2019aircraft, li2018remaining, ren2018bearing} discuss several methods for determining the RUL of an aircraft engine and bearing systems. Lu et al. \cite{lu2019aircraft} evaluate the usability of an online sequential extreme learning machine for prognostics. Li et al. \cite{li2018remaining} uses a deep convolution neural network and evaluates the performance on the C-MAPSS dataset. Ren et al. \cite{ren2018bearing} use an autoencoder for predicting the RUL of bearings. Despite recent progress, the DL-based approaches struggle as the complexity of the nonlinear systems increases. Additionally, the black-box nature of DL methods makes them undesirable in modeling real cyber-physical systems as DL-based approaches do not respect physical laws \cite{rai2020driven}. The black box nature and lack of explainability of DL models raise concerns for their usage in modeling real-life cyber-physical systems. However, there is work currently being conducted exploring the interpretability of neural networks for health monitoring and fault diagnosis. Wang et al. \cite{wang2022fully} have explored developing a fully interpretable neural network for locating resonance frequency bands in the NASA bearing datasets. They use a series of hidden layers to pick up typical signal processing features such as wavelet coefficients, square envelopes, square envelope spectrums, and sparsity measures. Secondly, Yang et al. \cite{yang2020interpreting} developed an attention mechanism to interpret the feature extraction process for bearing fault diagnosis. The output of the attention mechanism outlines the parts of the signal that the network is focusing on to make the prediction, which helps interpret the important features for fault diagnosis. Thirdly, Yang et al. \cite{yang2022interpretability} use neuron activation maximization to optimize the input signal and give clarity on the type of data required to make an optimal prediction. Yang et al. \cite{yang2022interpretability} also use saliency maps which determine the influence of different frequency segments on determining its health condition.
Additionally, DL-based approaches are "data-hungry." In many engineering applications, the dataset is often limited or sparse due to the high cost of collecting data via expensive simulations or physical experiments \cite{karpatne2017physics,jia2018physics, viswanathan2005fastplace}. A standard for selecting a DL technique, most effective at providing a prognosis, is not established as of the time of this publication--as it is highly dependent on the input data. 


Due to the limitations of both modeling techniques for RUL estimation, we pivot the attention to hybrid modeling--fusing PbM with DDM \cite{karpatne2017physics,singh2019pi,pillai2016hybrid,hanachi2017hybrid,young2017physically,kristjanpoller2014volatility}. Combining DDM methods with PbM methods can close the gap between partial physics models and complete models; therefore, reducing the time and effort required to build a complete system model and easing the need for accurately estimating system model parameters. Also, with part of the model determined and abstracted by PbM in the hybrid modeling approaches, DDM does not need to learn the complete nonlinear mapping--this improves the accuracy and reduces the amount of data required to train the model. There are three main approaches for achieving hybrid modeling:

(1) \textbf{Combination Approaches}: The output of the physics-based model is directly combined with the data-driven model either as an extra input (parallel combination) or as the only input to the data-driven model (sequential combination) \cite{ karpatne2017physics, jia2018physics, pillai2016hybrid}.  However, in this approach, since the physics model outcomes are merely an input to the data-driven model, the results are highly dependent on the accuracy of the physics model. Bolander et al. \cite{bolander2009physics} outlines the steps to determining the RUL of a bearing system and presents an approach where a DL model can be beneficial. However, the model relies on a sequential combination with a highly computational spall growth model, limiting the usefulness of the proposed approach. Another instance is 3D-PhysNet, introduced by Wang et al. \cite{wang20183d}. 3D-PhysNet is a deep variational encoder with a discriminator trained using the data from a finite element simulator to discover the behavior of deformable objects under external forces.

(2) \textbf{Constraint Approaches}: Physical information is employed as constraints to the learning process by incorporating the physical information in the loss function \cite{karpatne2017physics, muralidhar2018incorporating, stewart2017label} and considering the physical information as a constraint. Such hybrid modeling approaches aim to design a customized loss function to pay specific attention to different variables. Nonetheless, the existing works only propose simple constraints such as monotonicity and bounds \cite{muralidhar2018incorporating}. Elevating the efficacy of these approaches would require more complex constraints, but in practice, complex constraints invoke difficulties in the training procedure. An illustrative example of this approach is a paper by Pan et al. \cite{pan2018physics}. Pan et al. \cite{pan2018physics} introduce, PhysicsGAN, which implements a physics constraint on the training process. PhysicsGAN uses a GAN to improve the capabilities of deblurring images. The GAN is constrained by a physics model for the degradation process to guarantee that GAN produces images consistent with the observed inputs.

(3) \textbf{Embedding Approaches}: In embedding-based hybrid modeling approach, the data-driven model is used to replace the computationally expensive component in the physics-based model to speed up or enhance the physics-based solver \cite{long2018hybridnet, kim2019deep, liu2015model}.  Although these methods facilitate improved accuracy when the computational component is easy to split, the application domain of these methods is limited. Commonly, the physics models are complex and highly coupled, so these methods are difficult to apply widely. An example of this in the RUL estimation domain is discussed by Chao et al. \cite{chao2020fusing}. A unique way of fusing physics and DL to achieve a higher level of accuracy is introduced for the C-MAPSS dataset. The authors use physics-based performance models to infer unobservable model parameters related to a system's components' health. These parameters are subsequently combined with sensor readings and input to a deep neural network to generate a data-driven prognostics model with physics-augmented features. The method is computationally acceptable and outperforms standalone DDM.

We contribute to the hybrid modeling vein of system modeling research by introducing a GAN infused with physics using an embedding approach.
Although embedding approaches are harder to apply widely, we provide a framework to isolate a location for embedding the physics model for GANs. The embedded physics model is an impartial model for failure propagation. These failures are predicted using a series of parameters that are determined through computationally expensive modeling for simulated experimental data collection. Therefore, the model can be used without these parameters, and the DL model can map the input features of the data to estimate these parameters for predicting RUL. GANs \cite{goodfellow2014generative}, with less labeled data, outperform Deep Neural Networks (DNNs) \cite{deng2014deep} and Convolutional Neural Networks (CNNs) \cite{kim2017convolutional} in classification across various tasks; this shows promise for developing GANs for regression which has garnered significant traction \cite{rezagholizadeh2018semi, olmschenk2018generalizing, aggarwal2019benchmarking, nguyen2021fuzzy}. Nevertheless, most GANs suffer from a lack of stability; optimality is tricky because the slight changes in the hyperparameters can lead to the generator or discriminator dominating the game, making it challenging to train GANs. We draw attention to our prior work at the interface of fuzzy logic, and GANs \cite{nguyen2021fuzzy}, where we show an improvement in regression accuracy by introducing a differentiable fuzzy logic model in various locations of the GAN architecture. We build upon and leverage the regression injection technique \cite{nguyen2021fuzzy} which aggregates the fuzzy implications using a product aggregator for our proposed method in this paper. 

Nguyen et al. \cite{nguyen2021fuzzy}  suggest an exploration of more useful aggregations, so we propose using aggregators that bound the output of the network to abide by physical laws. Our modification to the FuzzyGAN model is inspired by previous work where physics is integrated with data-driven models to enhance the predictive capabilities--most of the current work clearly shows that the physics models are pivotal in guiding the data-driven predictions.



 The specific contributions of this paper are as follows:
\begin{itemize}
    \item We introduce a novel infusion of physics with GAN--specifically leveraging a FuzzyGAN framework: PhyzzyGAN. The outlined PhyzzyGAN model ensures physically realistic solutions by constraining the output of the entire model to fit the known physics. The infusion of physics is enabled by embedding the physics into the output of the DL model.
    \item We showcase the efficacy of the outlined model in health monitoring and prognostics tasks. Specifically, we show the utility of the outlined model on the NASA Bearing dataset. We evaluate the improvement of the predictive accuracy of the PhyzzyGAN architecture. The empirical results demonstrate the superior regression capabilities of PhyzzyGAN compared to the CGAN or GANs with fuzzy models thus ascertaining its specific utility in prognostics applications.
\end{itemize}

The following sections will discuss and evaluate the proposed novel infusion. Section 2 outlines an introduction to the proposed infusion method. Section 2 also develops a technical description of the infusion by explaining the chosen GAN architecture, outlining the mathematical theory behind fuzzy logic-based systems (and FuzzyGANs), and develops a rule base for the chosen GAN using physics models. Section 3 introduces the test problems and builds the different physics models used in the proposed framework. Section 3 also discusses the evaluation metrics. Section 4  reports the results. Finally, section 5 revisits the topics discussed in sections 2-4 and concludes the paper.

\section{Technical Description}

Our PhyzzyGAN model uses Conditional GAN (CGAN) \cite{mirza2014conditional} as the base framework and leverages a FuzzyGAN regression injection technique \cite{nguyen2021fuzzy} for the prediction. Next, the critical concepts related to CGAN and FuzzyGAN are briefly introduced. Additionally, technical details that are necessary for implementing PhyzzyGAN are outlined in this section. Finally, details about another DL model used to compare the performance of PhyzzyGAN are presented in this section.  

\subsection{Conditional GAN (CGAN)}
 CGAN \cite{mirza2014conditional} conditions both the generator and discriminator on the output, $y$. The generator receives the concatenation between the noise input, $z$, and output, $y$; while, the discriminator receives the concatenation between the input, $x$, and $y$. This architecture is considered because Aggarwal et al.
 \cite{aggarwal2019benchmarking} evaluated the efficacy of using CGAN for regression and noticed promising results over two metrics: Negative Log Predictive Density (NLPD) and Mean Absolute Error (MAE). Aggarwal et al.
 \cite{aggarwal2019benchmarking} modified the original GAN architecture so that the generator, $G(\cdot)$, receives two inputs: a noise input, $z$, and the true input, $x$. These inputs each feed into their respective fully-connected layer before eventually being concatenated and processed in the generator to predict $y$. 
 

The authors benchmark CGAN by comparing the NLPD and MAE of the architecture against modern machine learning architectures such as XGBoost, MDNs, and a DNN across seven datasets. Through their analysis, the authors show that CGAN is competitive with modern machine learning architectures for regression and noted that there is room for considerable improvement. FuzzyGAN \cite{nguyen2021fuzzy} and subsequently PhyzzyGAN showcase improvement in regression tasks over CGAN.

\subsection{FuzzyGAN}
FuzzyGAN adopts a GAN architecture and embeds a differentiable fuzzy logic (DFL) model to perform regression--DFL is a variant of fuzzy logic introduced by Van Krieken et al. \cite{van2020analyzing}. The following subsections provide technical preliminaries needed to understand the FuzzyGAN architecture.

\subsubsection{Differentiable Fuzzy Logic (DFL)}
DFL is the fuzzy logic model used in FuzzyGAN \cite{nguyen2021fuzzy}. FuzzyGAN provides a stability improvement of the GAN game by transforming the output of the generator from $\hat{y}$ to providing probabilities of existing abstract features. DFL takes the probabilities and uses functions to map those probabilities to $\hat{y}$. Fuzzy logic has shown experimental success in regression settings--the improvement was further shown by the upgrade on GANs for regression demonstrated by Nguyen et al. \cite{nguyen2021fuzzy}. DFL is an extension of fuzzy logic that includes strictly differentiable operators. DFL is composed of four main steps: calculating the t-norm, the t-conorm, the implications of those norms, then the aggregation of the implications \cite{van2020analyzing}.
Additionally, DFL defines new semantics extending traditional fuzzy logic semantics to vector embeddings. The vector embeddings can be further abstracted than the traditional semantics represented by words or phrases.  A structure in DFL consists of a domain of discourse and an embedded interpretation.

\begin{definition}
A DFL structure is a tuple $S = \langle O, \eta, \theta \rangle$, where $O$ is a finite but unbounded set called domain of discourse and every $o \in O$ is a d-dimensional vector, $\eta : P \times \mathbb{R}^W \rightarrow (O^m \rightarrow [0, 1])$ is an (embedded) interpretation, and $\theta \in \mathbb{R}^W$ are parameters. $\eta$ maps predicate symbols $p \in P$ with arity $m$ to a function of $m$ objects to a truth value $[0, 1]$. That is, $\eta(p, \theta) : O^m \rightarrow [0, 1]$. 
\end{definition}

That is, objects in DFL semantics are $d$-dimensional vectors of reals. The vectorized semantics are then used in the t-norm and t-conorm operators to resolve the network's truthness.

\subsubsection{Differentiable Fuzzy Implications}
\label{FuzzyImp}
The functions that are used to compute the implication of two truth values are called fuzzy implications.
\begin{definition}

 A fuzzy implication is a function $I : [0, 1]^2 \rightarrow [0, 1]$ so that for all $a, c \in [0, 1]$, $I(\cdot, c)$ is
decreasing, $I(a, \cdot)$ is increasing and for which $I(0, 0) = 1, I(1, 1) = 1$ and $I(1, 0) = 0.$
\end{definition}

Similarly, many algorithms satisfy the above definition that is differentiable implications. However, FuzzyGAN uses a product-based implication, the Reichenbach implication, as it performed best in the experiments conducted by  Van Krieken et al. \cite{van2020analyzing}:

\begin{equation}
    I_{RC}(T,S) = 1 - T + T \circ S,
\end{equation}

where $T$ is a t-norm and $S$ is a t-conorm. FuzzyGAN applies an additional sigmoidal function to the implication that yields a sigmoidal implication, which is also an implication \cite{van2020analyzing}. This additional sigmoidal function is used to counteract the corner behavior of the Reichenbach implication. This is if values are around $a=0$ and $c=0$, then a majority of the gradient decreases the antecedent--this behavior is undesirable; it does not accurately provide information for training FuzzyGAN. Equation \ref{sigI} outlines the modified implication operator used in FuzzyGAN.
\begin{equation}
\label{sigI}
    I(T,S) = \frac{(1 + e^{9 / 2}) \cdot \sigma(9(I_{RC}(T,S) - 1/ 2)) - 1}{e^{9/2} - 1}.
\end{equation}

Where $\sigma(\cdot)$ is the sigmoid function. 

\begin{equation}
    \sigma(x) = \frac{1}{1 + e^{-x}}.
\end{equation}

\subsubsection{Differentiable Fuzzy Aggregation}

Lastly, from predicate fuzzy logic, functions used to compute quantifiers like $\forall$, and $\exists$ are represented as aggregation operators.
 \begin{definition}
 \label{Defagg}
 An aggregation operator is a function $A : [0, 1]^n \rightarrow [0, 1]$ that is symmetric and increasing with respect to each dimension, and for which $A(0, ..., 0) = 0$ and $A(1, ..., 1) = 1$. A symmetric function is one
in which the output value is the same for every ordering of its arguments.
 \end{definition}

The aggregation operator combines all the implications to make the final prediction based on all the implied information. The aggregator used in FuzzyGAN is expressed in Equation \ref{agg}. The log-product aggregator from  Van Krieken et al. \cite{van2020analyzing} did not demonstrate the same level of success that they reported when \cite{nguyen2021fuzzy} applied the aggregator to FuzzyGAN. The best performing aggregator was the product aggregator for FuzzyGAN \cite{nguyen2021fuzzy}. PhyzzyGAN (the focus of this paper) modifies the aggregation operator by replacing it with a physics model.

\begin{equation}
    \label{agg}
    A(x_1,...,x_n) = \prod_{i=1}^n x_i.
\end{equation}

Where $x$ is the substitute for the implications, $I$.

\subsubsection{FuzzyGAN: Regression Injection}
\begin{figure}[H]
    \centering
    \includegraphics[width= 5in]{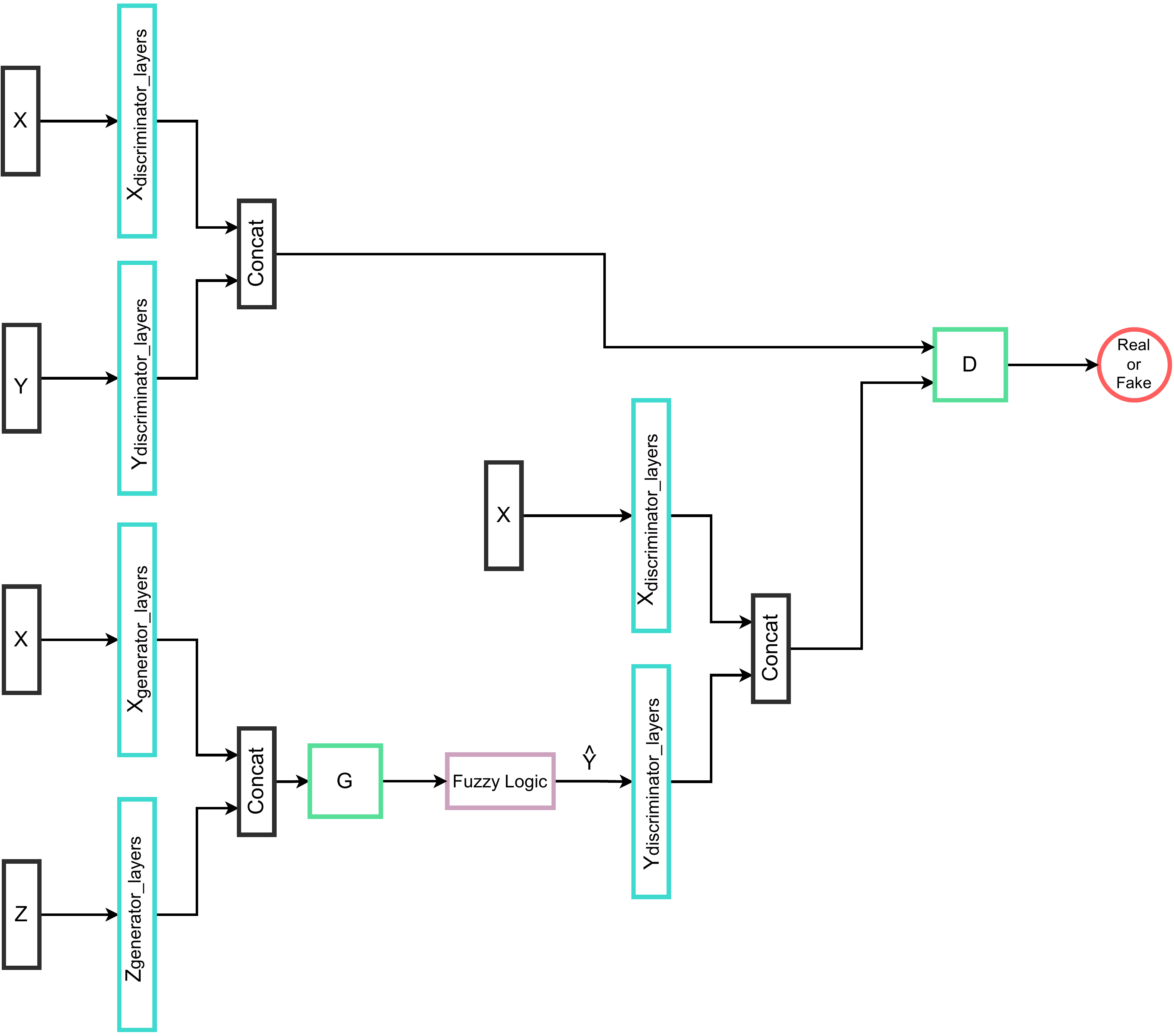}
    \caption{Flow chart for FuzzyGAN architecture with regression injection technique. FuzzyGAN uses a binary cross entropy loss function (Equation \ref{binarycross}) along with an adam optimizer \cite{kingma2014adam}. The number of epochs varied between problem sets so the rest of the hyperparameters are outlined in Appendix A. The two sets of inputs for $D$ are passed in an alternating way. The first pass is the concatenation of the processed $X$ and $Y$, but the second pass is the concatenation of the processed $X$ and $\hat{Y}$. This repeats until the networks are fully trained.}
    \label{fig:Reg_inj_1}
\end{figure}

With DFL developed, the regression injection technique can be outlined.  It injects the fuzzy logic to the output of the generator of the GAN that controls the prediction of $Y$. The main components of the FuzzyGAN architecture are shown in Figure \ref{fig:Reg_inj_1}.

\begin{equation}
    BCE Loss = -D(X,Y)log(D(X,G(X,Z))) + (D(X,Y)-1)log(1 – D(X,G(X,Z)))
    \label{binarycross}
\end{equation}

CGAN uses the concatenation of the input, $X$, along with noise, $Z$, in the generator, $G$, to predict the values of $Y$. $G$ is a neural network composed of convolutions that feed into a series of fully connected layers that outputs an $N$-dimensional vector. The network has to identify the number of abstract features where $N$ is the number of abstract features. These values represent a probability, $P$,  of those features existing. The $N$-dimensional output is then partitioned into four vectors: $a^{j \times 1}$, $b^{k \times 1}$, $c^{l \times 1}$, and $d^{m \times 1}$.  These four vectors represent the vector embeddings of the DFL model and can vary in size as long as $j + k + l + m = N$.  The size of the four vectors is a hyperparameter to be tuned. The process starts by testing the minimum size for each vector, so $j=1$, $k=1$, $l=1$, and $m=2$. Then, increase the size of each vector until an optimal size of the vectors is found. In this paper, the sizes of each vector are doubled simultaneously. The doubling provides a simple systematic method for exploring valid configurations with increasing vector sizes. $a^{j \times 1}$ and $b^{k \times 1}$ are combined in a conjunction, $T$, and $c^{l \times 1}$ and $d^{m \times 1}$ are combined in a disjunction, $S$. $T$ is then used as the antecedent, and $S$ is used as a consequent in a fuzzy implication, $I$. If the length of the longest vector is: $M$, the shape of $I$ should be: [batch size, $M$]. Each column, $M$, represents an implication, so the implications are aggregated over the columns using an aggregation operator, $A$, to create an output that is a vector the size of the batch size. The output of the fuzzy logic model is the prediction of $Y$, $\hat{Y}$. The input to the discriminator, $D$, is $X$ along with either $Y$ or $\hat{Y}$.  $D$ is also a neural network composed of convolutions that feed into a series of fully connected layers focused on learning to differentiate between $Y$ and $\hat{Y}$. The hyperparameters are outlined in Appendix A--the output of the generator is five and that is discussed Section \ref{datasetandevaluation}. FuzzyGAN demonstrated an improvement in CGAN's regression capabilities. However, it was noted that better results could be achieved by exploring different aggregation operators. PhyzzyGAN builds upon FuzzyGAN and allows for incorporating a physics model in place of an aggregation model and thus allows one to create hybrid models (Physics + data-driven).
\subsection{Proposed Model: Physics-Infused FuzzyGAN (PhyzzyGAN)}
The proposed architecture, PhyzzyGAN, modifies the FuzzyGAN architecture by replacing the aggregation operator with a physics model. By incorporating a physics model in this step the modified architecture aims to reduce the error in the regression metrics discussed in Section \ref{datasetandevaluation}.
\begin{figure}[H]
    \centering
    \includegraphics[width= 5in]{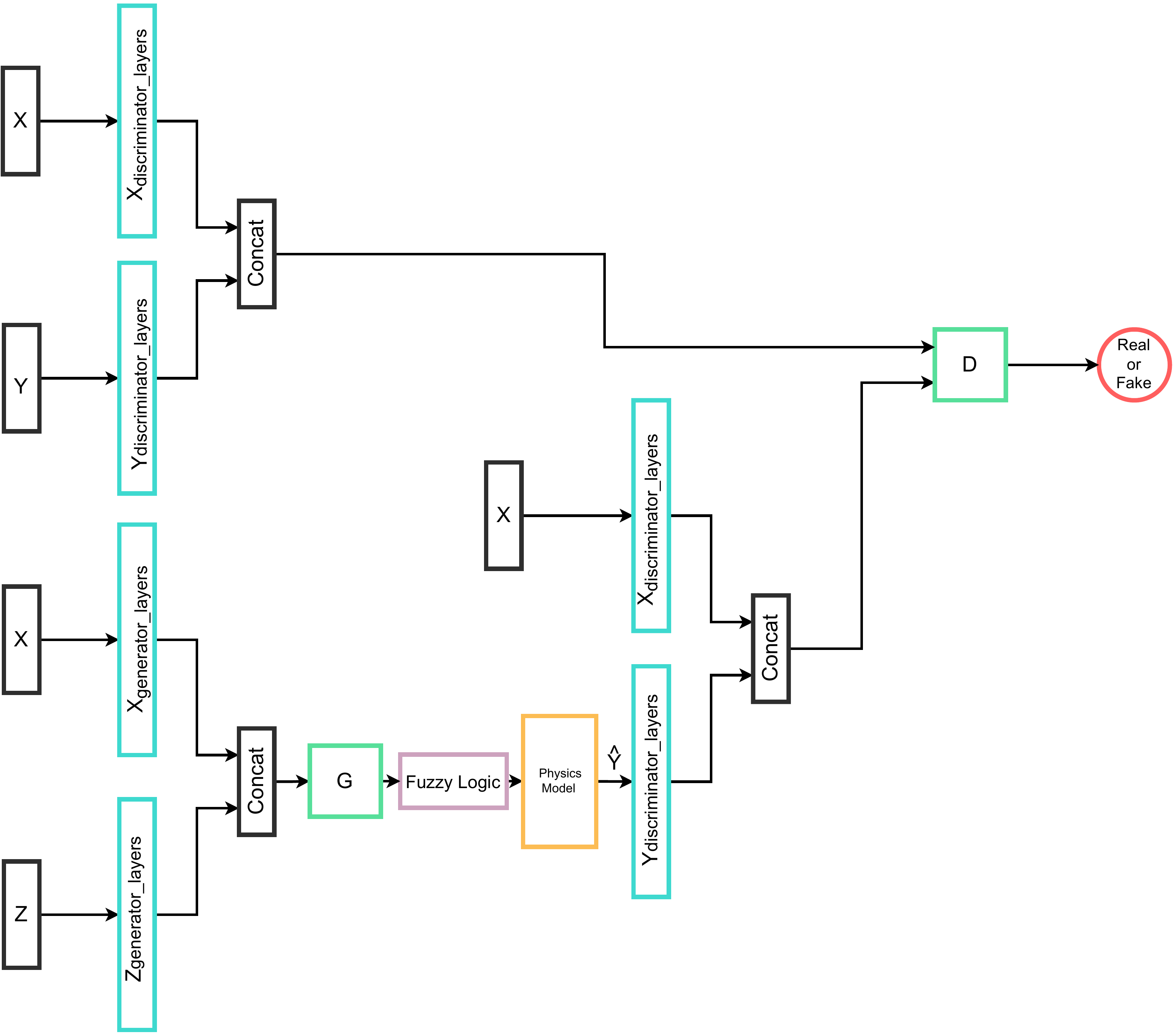}
    \caption{Flow chart for the proposed architecture: PhyzzyGAN. PhyzzyGAN uses a binary cross entropy loss function (Equation \ref{binarycross}) along with an adam optimizer \cite{kingma2014adam}. The number of epochs varied between problem sets so the rest of the hyperparameters are outlined in Appendix A. The two sets of inputs for $D$ are passed in an alternating way. The first pass is the concatenation of the processed $X$ and $Y$, but the second pass is the concatenation of the processed $X$ and $\hat{Y}$. This repeats until the networks are fully trained.}
    \label{fig:PhyzzyGAN}
\end{figure}
Figure \ref{fig:PhyzzyGAN} outlines the method of embedding the physics in the FuzzyGAN framework to create PhyzzyGAN. Like FuzzyGAN, $X$ is the input data, $Z$ is the noise, $Y$ is the output data, $G$ is the generator, and $D$ is the discriminator. However, PhyzzyGAN adds a physics model to the aggregation step in the fuzzy logic model in a FuzzyGAN. Therefore, unlike in FuzzyGAN, the fuzzy logic model block outputs the implications, $I_i$. The network learns features that help generate fuzzy implications about the missing aspects of the physics model. The physics model used as the aggregation operator is dataset-dependent. Therefore, we generated a different physics model for each dataset; they are outlined in Section \ref{Datasets}. This injection of physics is utilized to leverage the GAN and the fuzzy logic model. The number of outputs of the generator is carefully tuned to achieve optimal accuracy. The hyperparameters for the networks that make up PhyzzyGAN are outlined in Appendix A.

\section{Dataset and Evaluation Metrics}
\label{datasetandevaluation}
This section discusses the prognostics dataset used to evaluate the proposed technique and develop the physics model in the aggregation step of PhyzzyGAN. 
\label{Datasets}
\subsection{Bearings}

\subsubsection{Data Description}
The bearing dataset was compiled by Qiu et al. \cite{qiu2006wavelet}. The data collected was of four Rexnord ZA-2115 double row bearings installed on a shaft. The bearings were held at a constant rotation speed of 2000rpm. Additionally, a radial load of 6000 lbs was applied onto the shaft and bearing. All bearings were force lubricated. The type of data collected was accelerometer data, PCB 353B33 High Sensitivity Quartz ICP accelerometers were installed on the bearing housing two accelerometers for each bearing for the first dataset and one accelerometer each for the other two datasets.

\begin{figure}[H]
    \centering
    \includegraphics[width=5in]{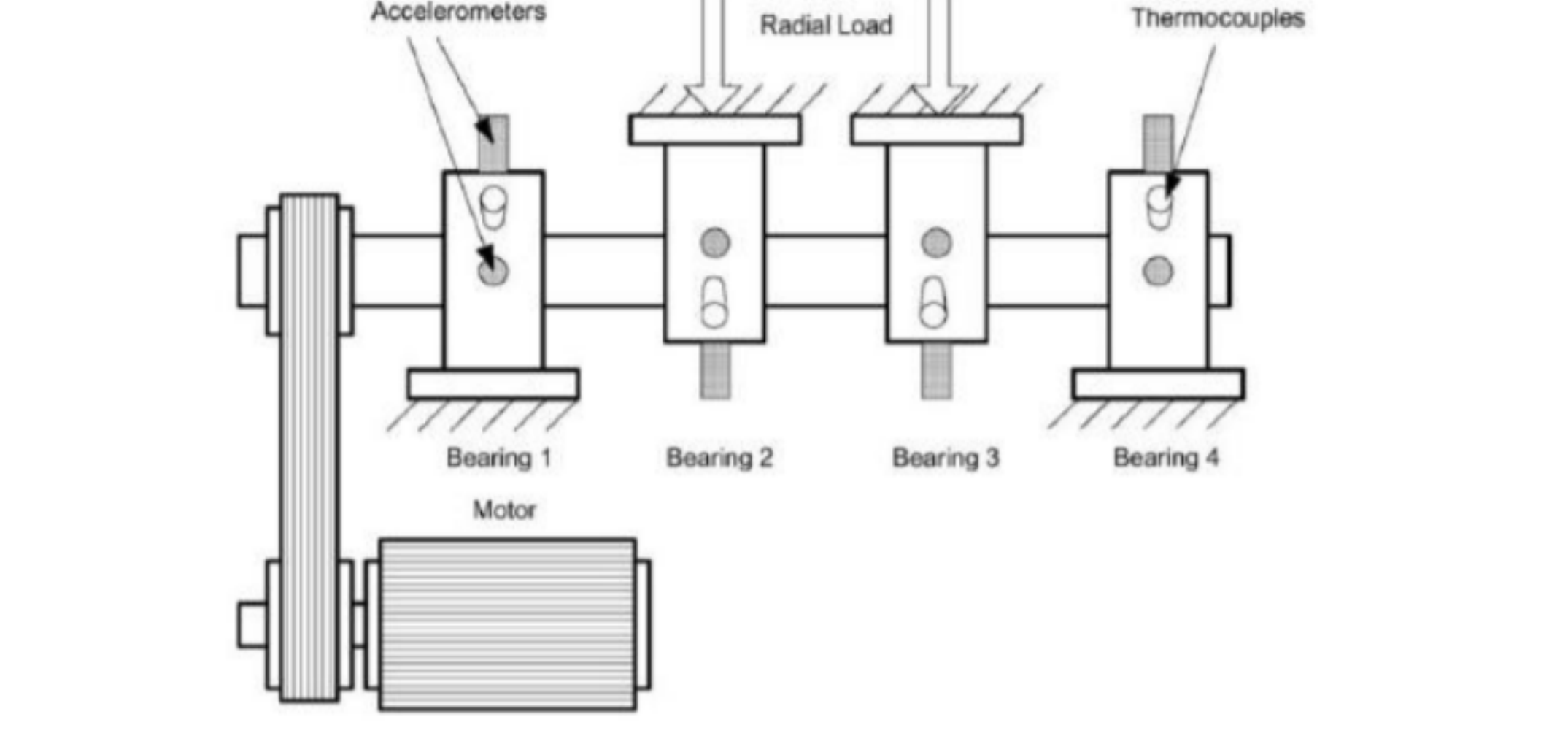}
    \caption{Schematic of the intelligent maintenance system (IMS) bearing rig test bed used in the Bearing dataset \cite{qiu2006wavelet}}
    \label{fig:bearing}
\end{figure}

Each dataset describes a test-to-failure experiment. Each dataset consists of individual 1-second vibration signal snapshots recorded at specific intervals. Each file consists of 20,480 points, with the sampling rate set at 20 kHz. Data collection was facilitated by NI DAQ Card 6062E \cite{qiu2006wavelet}. The experimental setup for data collection is shown in Figure \ref{fig:bearing}.

\subsubsection{Preprocessing}
The input features in the dataset are the acceleration of each bearing in terms of gravity ($g$) for a one-second interval. So there are $20,480\times4$ features, where 20,480 represents the sampled accelerations for one second, and 4 represents the number of bearings. First, we normalize the input data by using a z-score scaler. Then we create an output array by assigning an RUL value to each timestamp. The RUL is in terms of timestamps remaining–although the damage is nonlinear, the RUL is linear w.r.t. remaining timestamps. The bearing dataset is broken into three separate experiments, and the longest-lasting experiment contained 6324 timestamps. So the longest RUL value would be assigned to the first sample in that experiment. The RUL values for all three experiments were normalized by 6324; then, the normalized data is distributed into four separate training datasets: one for each experiment and a final one with the data from each experiment concatenated. The final dataset is split into a train and test set and is used for the evaluation of the architectures.

\subsubsection{Physics Model}
\label{BearingPhys}
The physics model used is derived from Sawalhi et al. \cite{sawalhi2017vibration} and is shown in equation \ref{BearingPhysics}. The bearings fail due to spalls growing on either the inner or outer race; thus, the core idea behind the model is that we use signal processing techniques to identify early weak faults and then calculate the spall width at that time. Afterward, we propagate it forward in time using an exponential growth model to determine the remaining useful life (RUL).
\begin{equation}
    l_o = \frac{\pi f_r(D_p^2-D_b^2) }{D_pf_s}sp + \sqrt{D_b\delta + \delta^2}
    \label{BearingPhysics}
\end{equation}

Where $l_o$ is the spall width, $D_p$ is the pitch diameter, $D_b$ is the ball diameter,  $f_r$ is the shaft rotational speed, $f_s$ is the sampling frequency, $sp$ is the number of samples between the entry and the impact, and $\delta$ is the fault depth.

The values of the parameters $D_p$, $D_b$, $f_r$, $f_s$ are known apriori \cite{qiu2006wavelet}; and the values of the parameters $sp$ and $\delta$ need to be determined from the data.
Chen and Kurfess \cite{CHEN201916} introduce a DDM technique for determining the entry and impact points of the spall--we adopt this method to acquire the missing component of our physics model. 

Chen and Kurfess \cite{CHEN201916} show that at the entry point, the change in the vibration signal decreases slowly to a minimum level before a major change. However, the raw signal has a low signal-to-noise ratio which makes it difficult to locate this point. When the entry signal is exposed to very strong high frequency background noises and disturbances, it is recommended that a varaiational mode decomposition (VMD) \cite{dragomiretskiy2013variational} is to be applied.

\begin{figure}[H]
    \centering
    \includegraphics[width=5in]{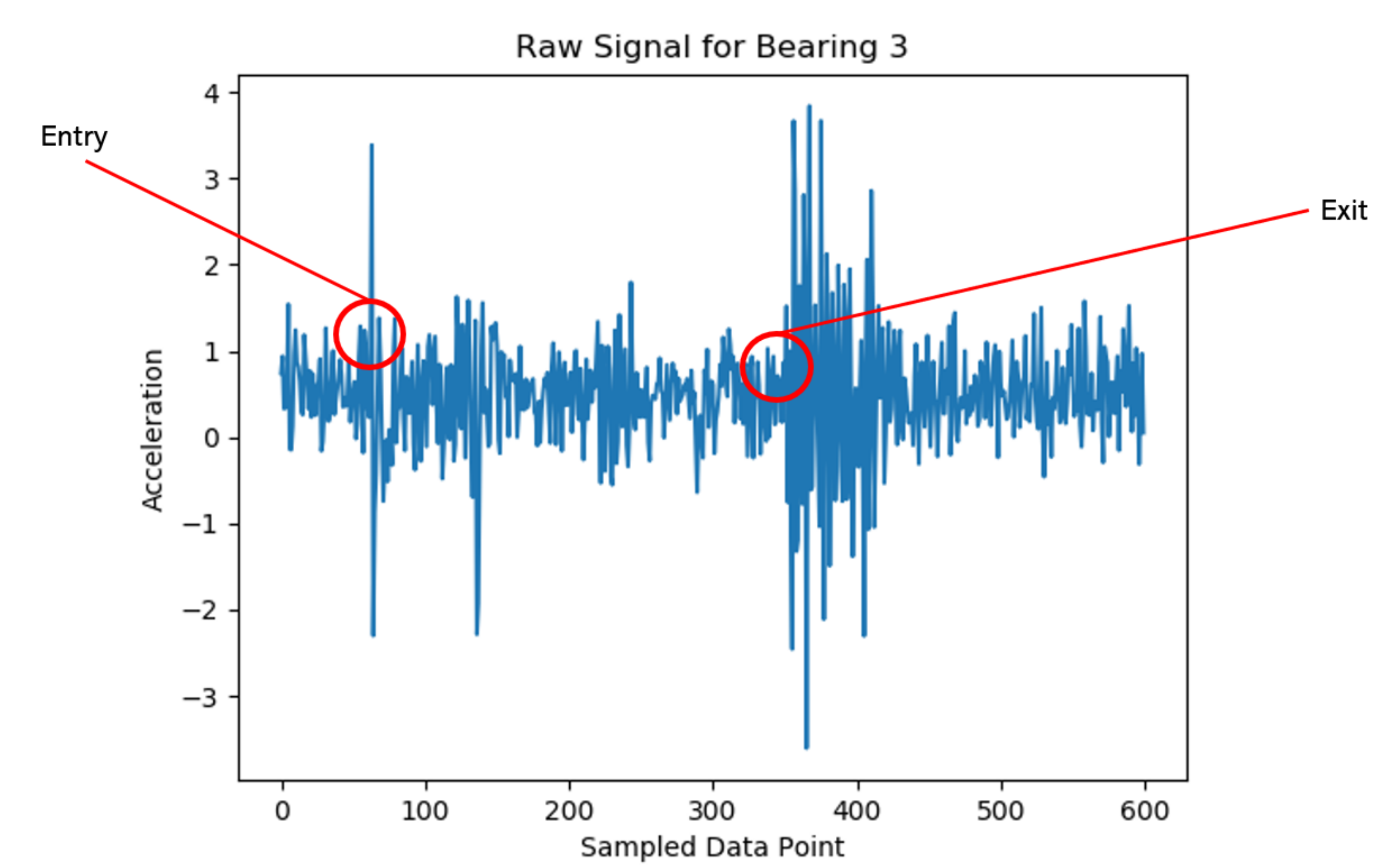}
    \caption{The first 600 unprocessed vibration signal data points of the 20,480 points sampled from the third bearing from file 2004.04.18.02.42.55 from the third test folder. The acceleration is measured in $g$}
    \label{fig:x}
\end{figure}

\begin{figure}[H]
    \centering
    \includegraphics[width=5in]{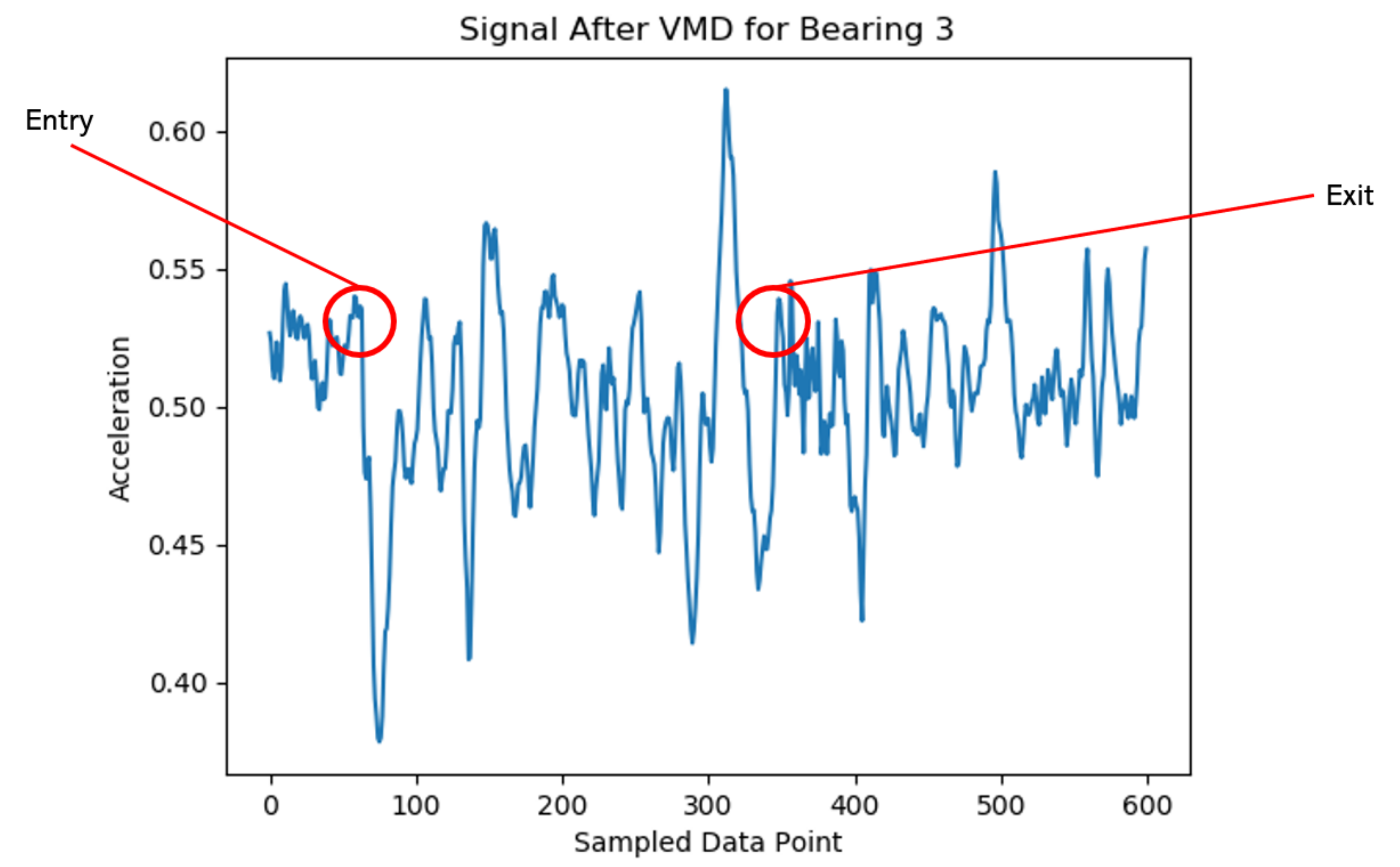}
    \caption{The first 600 vibration signal data points of the 20,480 points after the Variational Mode Decomposition was applied. This data comes from the first bearing from file 2004.04.18.02.42.55 from the third test folder. The acceleration is measured in $g$}
    \label{fig:VMD}
\end{figure}

In Figure \ref{fig:x}, an example of the time series of the vibration data is shown. The red circles demonstrate the difficulty of identifying the entry and exit points of the bearing from the raw data. So, for the entry point, the data is conditioned using a VMD with three modes to remove high-frequency noise. The post-processed initial data using the VMD method is shown in Figure \ref{fig:VMD}.

Figure \ref{fig:VMD} shows how removing the noise smooths the data and reduces the difficulty of finding the entry and exit points. To automate the location of the entry point, the noise-reduced signal is integrated using cumulative trapezoid numerical integration to develop a velocity model. A fourth-order empirical model is used to fit the data and represent the velocity model as a function of the rotation frequency and time. The accompanying equation is shown below.

\begin{equation}
    v = \psi(p_1\psi^4t^4 + p_2\psi^3t^3 + p_3\psi^2t^2 + p_4\psi t + p_5  )
    \label{velocity}
\end{equation}

\begin{figure}[H]
    \centering
    \includegraphics[width=5in]{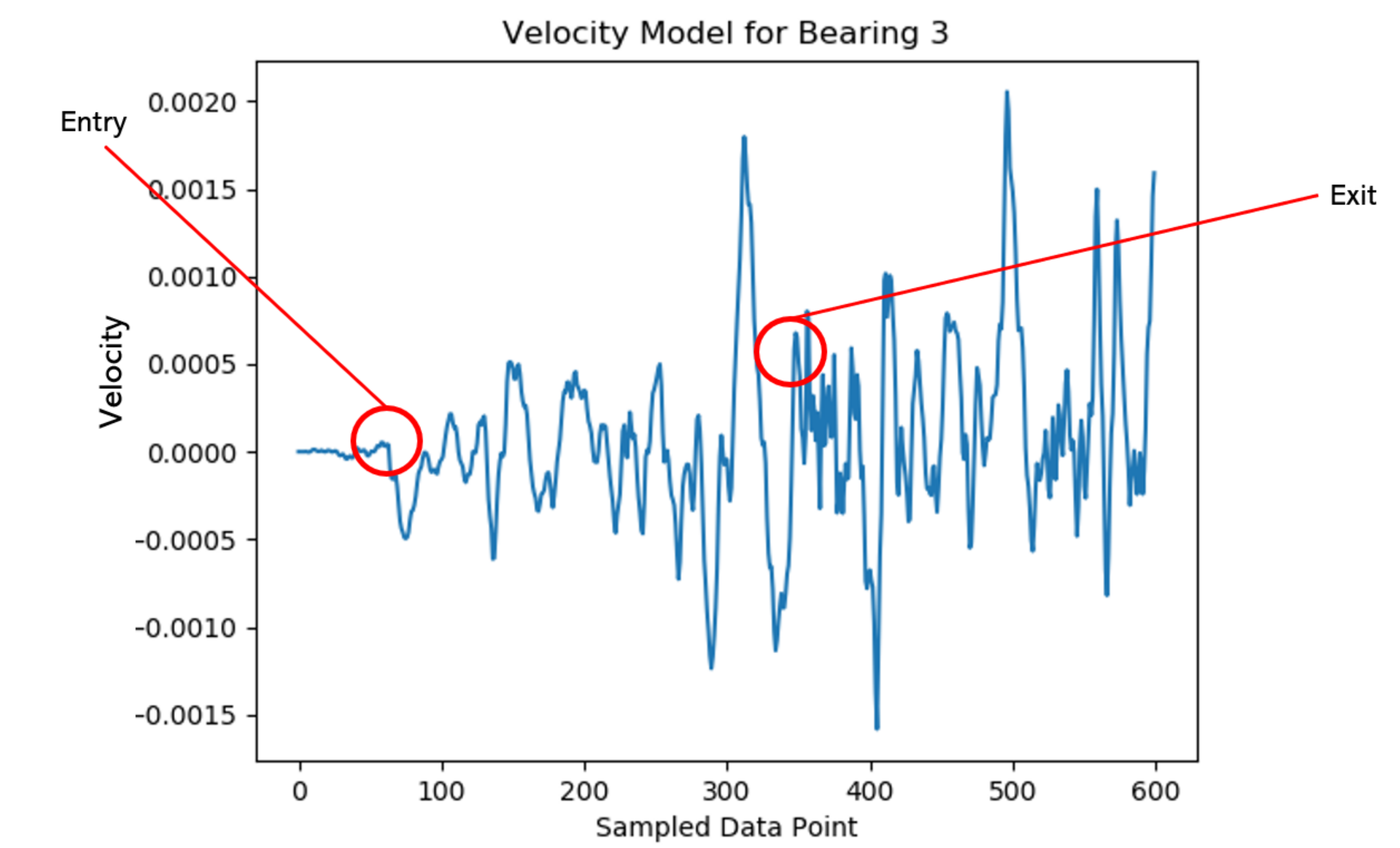}
    \caption{Velocity model of the first 600 vibration signal data points of the 20,480 points after the VMD was integrated. This data comes from the third bearing from file 2004.04.18.02.42.55 from the third test folder. The velocity is measured in $m/s$}
    \label{fig:integrated}
\end{figure}

Where $p_1$, $p_2$, $p_3$, $p_4$, and $p_5$ are the coefficients of the polynomial and $t$ is time. $\psi$ is $n/1000$rpm. For the dataset used $n=2000$rpm and $\psi=2$. The velocity model is shown in Figure \ref{fig:integrated}.

The velocity function (equation \ref{velocity}) is then symbolically differentiated to create an empirical model for acceleration:

\begin{equation}
        a = \psi(4p_1\psi^4t^3 + 3p_2\psi^3t^2 + 2p_3\psi^2t + p_4\psi).
         \label{acceleration}
\end{equation}

Next, the time $t_m$ when the acceleration reaches a local minimum and the minimum acceleration $a_{min}$ is determined. Additionally,  the velocity at $t_m$ is determined and set equal to $k$. Then using equation \ref{entry}, the entry point is determined.

\begin{equation}
    Entry = t_m + \frac{k\psi}{a_{min}}
    \label{entry}
\end{equation}

\begin{figure}[H]
    \centering
    \includegraphics[width=5in]{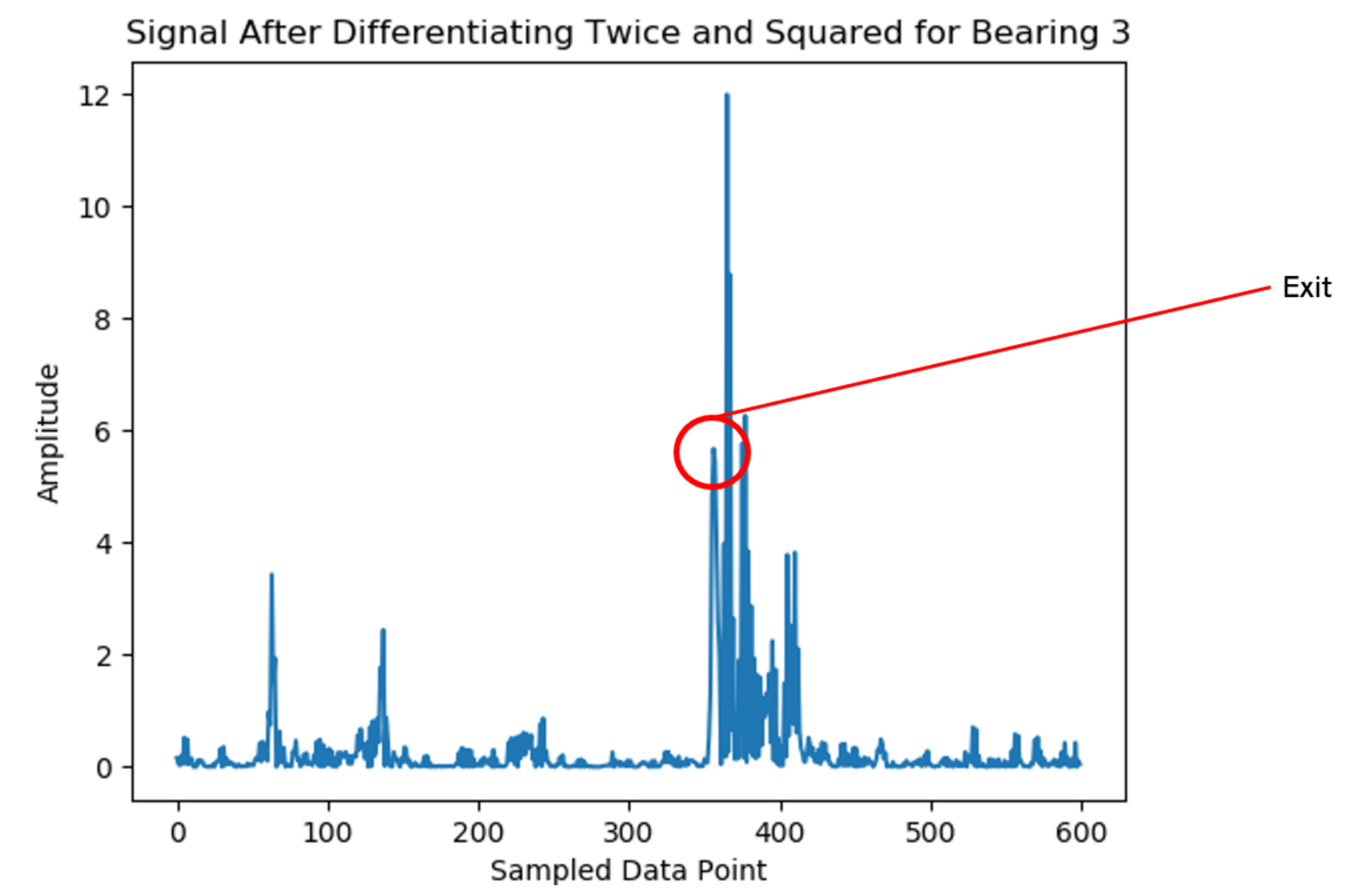}
    \caption{The first 600 vibration signal data points of the 20,480 points after the signal was twice differentiated and squared. This data comes from the first bearing from file 2004.04.18.02.42.55 from the third test folder.}
    \label{fig:final}
\end{figure}

The impact point is determined by differentiating the raw signal twice then squaring the signal. After the entry point where all criteria for determining a peak are achieved, the first sample is considered the impact point. The criterion for determining a peak is any point at $5\%$ of the maximum value of that particular set. This criterion causes an undershoot of the exit point, which is undesirable. However, this value was tuned to achieve the best performance. A greater threshold produced higher $sp$ values, but the higher $sp$ values yielded less accurate predictions from PhyzzyGAN. Therefore, the $5\%$ of the maximum value yielded the best results for the overall framework. Figure \ref{fig:final}, shows the plot after transformation.

The difference between the impact point and the entry point is used to calculate $sp$. 


The concatenated data is iterated upon to extract $sp$ for each index to determine the RUL.  Then $sp$ is used in the physics model in equation \ref{BearingPhysics} to calculate $l_o$. Lastly, $l_o$ is used in an exponential growth equation to calculate $RUL$.

\begin{equation}
    RUL = \frac{\ln{(\prod_{i=1}^M{(I_i)}l_{max}/l_o)}}{\ln{(1+r)}}
\end{equation}

Where $l_{max}$ is the width of the spall at failure, which uses $sp=600$, $r$, which is the rate of growth, that is preset as 0.001, and $I_i$ represents the implications of the fuzzy logic model. The physics model is weighted using the sum of products of implications. The implications represent the confidence in the physics model for making the prediction. We normalize the output by dividing by the maximum amount of time intervals set to a value of 6324. If the physics model predicts a value too high or too low, the network will learn to modify that value through fuzzy implications.

\subsubsection{Fuzzy Logic}
The number of implications was tested for powers of 2. We started with two implications and found that 16 were optimal for this application--it outperformed configurations with more or fewer implications. The values for $j$, $k$, $l$, and $m$ were explored where $j=k=l=m/2$. So our final configuration was $j=k=l=8$ and $m=16$.




\subsection{Evaluation Metrics}
Three different evaluation metrics are used to measure the approach's efficacy in improving the regression capabilities. The three main metrics are:  mean absolute error (MAE), mean squared error (MSE), and the C-MAPSS score function (S).

\textit{Mean Absolute Error (MAE)}

The MAE is the absolute value of the difference between the true output and the predicted output averaged over the number of samples. MAE gives an even measure of how well our model is performing. The absolute value means that this evaluation metric is not sensitive to outliers. The equation for MAE is shown in equation \ref{MAE}.

\begin{equation}
    MAE = \frac{\sum_{i=1}^n|y_i-\hat{y}_i|}{n}
    \label{MAE}
\end{equation}

Where our outputs are scaled between 0 and 1 using a min-max values because the fuzzy logic model outputs a value between 0 and 1.

\textit{Mean Squared Error (MSE)}

The MSE is the square of the difference between the true output and the predicted output averaged over the number of samples. The MSE is excellent for ensuring that our trained model has no outlier predictions with huge errors. The MSE puts more considerable weight on these errors due to the squaring part of the function. The following equation is used for computing MSE:

\begin{equation}
    MSE = \frac{\sum_{i=1}^n(y_i-\hat{y}_i)^2}{n}
\end{equation}

Where our outputs are scaled between 0 and 1 using a min-max values because the fuzzy logic model outputs a value between 0 and 1.





\section{Results} 


After infusing a physics model in the aggregation operator, the following results outline the performance upgrade that evaluates the fuzzy implications over two metrics: mean absolute error (MAE) and mean squared error (MSE). We compare the performance of the Physics-Infused FuzzyGAN (PhyzzyGAN) model with three contemporary CGAN models to show how varying embedding levels affect the regression capabilities of these respective architectures. The three models used for comparative analysis are the conventional CGAN, FuzzyGAN, and PhysiCGAN. CGAN is used to measure the base performance of this algorithm. Then FuzzyGAN is a regression injected CGAN used to measure the performance upgrade using fuzzy logic. PhysiCGAN is a proposed design that fuses the base CGAN and a physics model without fuzzy logic. Lastly, PhyzzyGAN combines the regression injected fuzzy logic GAN and the physics model. The empirical results demonstrate the improvement in the regression capabilities of the GAN architecture by embedding both a fuzzy logic model and a physics model. The potency of the proposed PhyzzyGAN architecture is demonstrated over two real-world problem sets, Bearing and C-MAPSS. The results for the bearing dataset establish the efficacy of the proposed pipeline even in the presence of an insufficient partial-physics model. In this case, the model can leverage signal processing techniques to provide a value for a vital missing component. 
In comparison, C-MAPSS measures the performance enhancement in the presence of a complete physics model.

\subsection{Bearing}
As mentioned in Section \ref{BearingPhys}, due to the lack of sufficient levels of information abstraction in the partial physics model, hybrid models using different signal processing techniques outlined in Section \ref{BearingPhys} were constructed.  

\begin{table}[H]
    \centering
    \begin{tabular}{|c|c|c|}
    \hline
         Network & MAE & MSE\\
         \hline
         CGAN & $0.2733 \pm 0.003327$ & $0.08673 \pm 0.0007969$ \\
         \hline
         FuzzyGAN & $0.2585 \pm 0.001998$ & $0.08728 \pm 0.0004388$ \\
         \hline

         \textbf{PhyzzyGAN} & \textbf{$0.2284 \pm$} \textbf{$0.000$} & \textbf{$0.07586 \pm$} \textbf{ $0.000$}\\
         \hline
    \end{tabular}
    \caption{Comparison between the three architectures for the Bearing dataset}
    \label{tab:Bearing_Res}
\end{table}

First, we will compare FuzzyGAN to CGAN because it shows a higher value for the MSE metric. The results show that adding a physics-inspired model or a fuzzy logic system, separately, helps improve the base model's prediction accuracy; however, the higher MSE points to this model producing either more outliers or more extreme outliers. Due to the nature of the data, it is difficult to predict the RUL from the bearing dataset because a fault may not have occurred.  The physics of the system does not drive FuzzyGAN, so the MSE is similar to CGAN's MSE results. These issues are alleviated when we create the PhyzzyGAN framework--Table \ref{tab:Bearing_Res} shows that PhyzzyGAN demonstrates the most significant performance improvement. There is a $16.43\%$ decrease in the MAE and a $12.53\%$ decrease in the MSE from CGAN. The physics model aids in improving the accuracy of the hybrid model by accounting for nonlinearities in the mapping, while the fuzzy logic model is included to provide a confidence value for that physics model.  The addition of a fuzzy logic model helps reduce the outliers for cases where the physics model does not apply (i.e., a fault is not present), and the fuzzy logic model can have a more decisive influence on the network's output.

In conclusion, the results show an improvement; however, the aggregation of the implications abides by the symmetry constraint for aggregation operators; future work could explore an entanglement of the physics model and fuzzy logic model that breaks this constraint to see if a better performance upgrade can be achieved. A better model would reduce the output outliers and account for the lack of applicability when a fault does not exist. Reduced outliers in the output would result in a lower MSE. In future explorations, we can test models and emphasize minimizing MSE to maximize the potential of the PhyzzyGAN model.

\section{Conclusion}
This paper presents a novel embedding of a physics model in the aggregation module of the injected fuzzy logic model to improve GAN architectures for regression tasks that are important in prognostics applications. We show that the fuzzy logic coupled with the physics-based model help to improve the regression capabilities of PhyzzyGAN and thus can be used for predicting RUL in prognostics application. The paper measures performance across two separate metrics for the bearing dataset. Each is used to assess different behaviors of the network. Across MAE and MSE, the bearing dataset showed a $16.43\%$ and a $12.53\%$ improvement. The performance enhancement can be attributed to the superior quality of physics models.  The paper focuses on an innovative combination of fuzzy logic and GANs for prognostics applications, but future research should improve the discussed hybrid model to compete with hybrid models leveraging other types of deep learning modules. Improving the physics models used and improving the embedding of the fuzzy implications into the physics models could yield more promising results.

Moreover, as physics models improve, the embedding of the fuzzy logic model will need to be revisited. We show that the fuzzy logic model loses value when the physics model is accurate enough. It reduces the number of outliers, but if there are not many outliers in the prediction, then the impact of the fuzzy logic model will diminish.  The fuzzy logic model could either be removed entirely or applied in other ways when this point is achieved.  One way could be including the fuzzy logic model in the discriminator to aid in differentiating between real and generated data and not entangle the fuzzy logic model and physics model. 

These conclusions highlight the fusion of PbM and DDM to improve the understanding and modeling of complex systems. However, we believe there are many avenues for improving the hybrid model results. The fidelity of the physics model plays a significant role in the accuracy of the hybrid model. If the relationship between the incomplete physics model and the actual model is more complex than the relationship between the inputs to the actual model, then adding partial physics will not improve performance. Determining a method to measure the quality of the physics model can improve our model decision-making, improving the efficacy of the hybrid model. 

We propose pivoting the focus of researchers in exploring other methods of introducing physics into the GAN architecture. We believe there are other physics models to evaluate and many areas to inject a physics model. The injection point may vary between datasets. A grasp of this underlying decision-making could lead to a more comprehensive understanding of hybrid modeling.
\bibliographystyle{elsarticle-num}
\bibliography{References}

\begin{thebibliography}{10}
\expandafter\ifx\csname url\endcsname\relax
  \def\url#1{\texttt{#1}}\fi
\expandafter\ifx\csname urlprefix\endcsname\relax\def\urlprefix{URL }\fi
\expandafter\ifx\csname href\endcsname\relax
  \def\href#1#2{#2} \def\path#1{#1}\fi

\bibitem{parlos2000multi}
A.~G. Parlos, O.~T. Rais, A.~F. Atiya, Multi-step-ahead prediction using
  dynamic recurrent neural networks, Neural networks 13~(7) (2000) 765--786.

\bibitem{qiu2002damage}
J.~Qiu, B.~B. Seth, S.~Y. Liang, C.~Zhang, Damage mechanics approach for
  bearing lifetime prognostics, Mechanical systems and signal processing 16~(5)
  (2002) 817--829.

\bibitem{cubillo2016review}
A.~Cubillo, S.~Perinpanayagam, M.~Esperon-Miguez, A review of physics-based
  models in prognostics: Application to gears and bearings of rotating
  machinery, Advances in Mechanical Engineering 8~(8) (2016) 1687814016664660.

\bibitem{gao2019wiener}
Z.~Gao, X.~Yin, B.~Zhang, M.~Chen, B.~Li, A wiener process--based remaining
  life prediction method for light-emitting diode driving power in rail vehicle
  carriage, Advances in Mechanical Engineering 11~(3) (2019) 1687814019832215.

\bibitem{sun2021improved}
B.~Sun, Y.~Li, Z.~Wang, Y.~Ren, Q.~Feng, D.~Yang, An improved inverse gaussian
  process with random effects and measurement errors for rul prediction of
  hydraulic piston pump, Measurement 173 (2021) 108604.

\bibitem{dong2019data}
G.~Dong, F.~Yang, Z.~Wei, J.~Wei, K.-L. Tsui, Data-driven battery health
  prognosis using adaptive brownian motion model, IEEE Transactions on
  Industrial Informatics 16~(7) (2019) 4736--4746.

\bibitem{hite1993algorithm}
S.~W. Hite~III, An algorithm for determination of bearing health through
  automated vibration monitoring, Tech. rep., ARNOLD ENGINEERING DEVELOPMENT
  CENTER ARNOLD AFB TN (1993).

\bibitem{wang2016size}
W.~Wang, N.~Sawalhi, A.~Becker, Size estimation for naturally occurring bearing
  faults using synchronous averaging of vibration signals, Journal of Vibration
  and Acoustics 138~(5) (2016).

\bibitem{sawalhi2017vibration}
N.~Sawalhi, W.~Wang, A.~Becker, Vibration signal processing for spall size
  estimation in rolling element bearings using autoregressive inverse
  filtration combined with bearing signal synchronous averaging, Advances in
  Mechanical Engineering 9~(5) (2017) 1687814017703007.

\bibitem{soons2020predicting}
Y.~Soons, R.~Dijkman, M.~Jilderda, W.~Duivesteijn, Predicting remaining useful
  life with similarity-based priors, in: International Symposium on Intelligent
  Data Analysis, Springer, 2020, pp. 483--495.

\bibitem{huang2019improved}
C.-G. Huang, H.-Z. Huang, W.~Peng, T.~Huang, Improved trajectory
  similarity-based approach for turbofan engine prognostics, Journal of
  Mechanical Science and Technology 33~(10) (2019) 4877--4890.

\bibitem{pinheiro2015learning}
P.~O. Pinheiro, R.~Collobert, P.~Doll{\'a}r, Learning to segment object
  candidates, in: Advances in Neural Information Processing Systems, 2015, pp.
  1990--1998.

\bibitem{kumar2016ask}
A.~Kumar, O.~Irsoy, P.~Ondruska, M.~Iyyer, J.~Bradbury, I.~Gulrajani, V.~Zhong,
  R.~Paulus, R.~Socher, Ask me anything: Dynamic memory networks for natural
  language processing, in: International Conference on Machine Learning, 2016,
  pp. 1378--1387.

\bibitem{lu2019aircraft}
F.~Lu, J.~Wu, J.~Huang, X.~Qiu, Aircraft engine degradation prognostics based
  on logistic regression and novel os-elm algorithm, Aerospace Science and
  Technology 84 (2019) 661--671.

\bibitem{li2018remaining}
X.~Li, Q.~Ding, J.-Q. Sun, Remaining useful life estimation in prognostics
  using deep convolution neural networks, Reliability Engineering \& System
  Safety 172 (2018) 1--11.

\bibitem{ren2018bearing}
L.~Ren, Y.~Sun, J.~Cui, L.~Zhang, Bearing remaining useful life prediction
  based on deep autoencoder and deep neural networks, Journal of Manufacturing
  Systems 48 (2018) 71--77.

\bibitem{rai2020driven}
R.~Rai, C.~K. Sahu, Driven by data or derived through physics? a review of
  hybrid physics guided machine learning techniques with cyber-physical system
  (cps) focus, IEEE Access (2020).

\bibitem{wang2022fully}
D.~Wang, Y.~Chen, C.~Shen, J.~Zhong, Z.~Peng, C.~Li, Fully interpretable neural
  network for locating resonance frequency bands for machine condition
  monitoring, Mechanical Systems and Signal Processing 168 (2022) 108673.

\bibitem{yang2020interpreting}
Z.-b. Yang, J.-p. Zhang, Z.-b. Zhao, Z.~Zhai, X.-f. Chen, Interpreting network
  knowledge with attention mechanism for bearing fault diagnosis, Applied Soft
  Computing 97 (2020) 106829.

\bibitem{yang2022interpretability}
H.~Yang, X.~Li, W.~Zhang, Interpretability of deep convolutional neural
  networks on rolling bearing fault diagnosis, Measurement Science and
  Technology 33~(5) (2022) 055005.

\bibitem{karpatne2017physics}
A.~Karpatne, W.~Watkins, J.~Read, V.~Kumar, Physics-guided neural networks
  (pgnn): An application in lake temperature modeling, arXiv preprint
  arXiv:1710.11431 (2017).

\bibitem{jia2018physics}
X.~Jia, A.~Karpatne, J.~Willard, M.~Steinbach, J.~Read, P.~C. Hanson, H.~A.
  Dugan, V.~Kumar, Physics guided recurrent neural networks for modeling
  dynamical systems: Application to monitoring water temperature and quality in
  lakes, arXiv preprint arXiv:1810.02880 (2018).

\bibitem{viswanathan2005fastplace}
N.~Viswanathan, C.-N. Chu, Fastplace: efficient analytical placement using cell
  shifting, iterative local refinement, and a hybrid net model, IEEE
  Transactions on Computer-Aided Design of Integrated Circuits and Systems
  24~(5) (2005) 722--733.

\bibitem{singh2019pi}
S.~K. Singh, R.~Yang, A.~Behjat, R.~Rai, S.~Chowdhury, I.~Matei, Pi-lstm:
  Physics-infused long short-term memory network, in: 2019 18th IEEE
  International Conference on Machine Learning and Applications (ICMLA), IEEE,
  2019, pp. 34--41.

\bibitem{pillai2016hybrid}
P.~Pillai, A.~Kaushik, S.~Bhavikatti, A.~Roy, V.~Kumar, A hybrid approach for
  fusing physics and data for failure prediction, International Journal of
  Prognostics and Health Management 7~(025) (2016) 1--12.

\bibitem{hanachi2017hybrid}
H.~Hanachi, W.~Yu, I.~Kim, C.~Mechefske., Hybrid physics-based and data-driven
  phm (2017).

\bibitem{young2017physically}
C.-C. Young, W.-C. Liu, M.-C. Wu, A physically based and machine learning
  hybrid approach for accurate rainfall-runoff modeling during extreme typhoon
  events, Applied Soft Computing 53 (2017) 205--216.

\bibitem{kristjanpoller2014volatility}
W.~Kristjanpoller, A.~Fadic, M.~C. Minutolo, Volatility forecast using hybrid
  neural network models, Expert Systems with Applications 41~(5) (2014)
  2437--2442.

\bibitem{bolander2009physics}
N.~Bolander, H.~Qiu, N.~Eklund, E.~Hindle, T.~Rosenfeld, Physics-based
  remaining useful life prediction for aircraft engine bearing prognosis, in:
  Annual Conference of the PHM Society, Vol.~1, 2009.

\bibitem{wang20183d}
Z.~Wang, S.~Rosa, B.~Yang, S.~Wang, N.~Trigoni, A.~Markham, 3d-physnet:
  Learning the intuitive physics of non-rigid object deformations, arXiv
  preprint arXiv:1805.00328 (2018).

\bibitem{muralidhar2018incorporating}
N.~Muralidhar, M.~R. Islam, M.~Marwah, A.~Karpatne, N.~Ramakrishnan,
  Incorporating prior domain knowledge into deep neural networks, in: 2018 IEEE
  International Conference on Big Data (Big Data), IEEE, 2018, pp. 36--45.

\bibitem{stewart2017label}
R.~Stewart, S.~Ermon, Label-free supervision of neural networks with physics
  and domain knowledge, in: Thirty-First AAAI Conference on Artificial
  Intelligence, 2017.

\bibitem{pan2018physics}
J.~Pan, J.~Dong, Y.~Liu, J.~Zhang, J.~Ren, J.~Tang, Y.-W. Tai, M.-H. Yang,
  Physics-based generative adversarial models for image restoration and beyond,
  arXiv preprint arXiv:1808.00605 (2018).

\bibitem{long2018hybridnet}
Y.~Long, X.~She, S.~Mukhopadhyay, Hybridnet: integrating model-based and
  data-driven learning to predict evolution of dynamical systems, arXiv
  preprint arXiv:1806.07439 (2018).

\bibitem{kim2019deep}
B.~Kim, V.~C. Azevedo, N.~Thuerey, T.~Kim, M.~Gross, B.~Solenthaler, Deep
  fluids: A generative network for parameterized fluid simulations, in:
  Computer Graphics Forum, Vol.~38, Wiley Online Library, 2019, pp. 59--70.

\bibitem{liu2015model}
B.~Liu, G.~Mason, J.~Hodgson, Y.~Tong, M.~Desbrun, Model-reduced variational
  fluid simulation, ACM Transactions on Graphics (TOG) 34~(6) (2015) 244.

\bibitem{chao2020fusing}
M.~A. Chao, C.~Kulkarni, K.~Goebel, O.~Fink, Fusing physics-based and deep
  learning models for prognostics, arXiv preprint arXiv:2003.00732 (2020).

\bibitem{goodfellow2014generative}
I.~Goodfellow, J.~Pouget-Abadie, M.~Mirza, B.~Xu, D.~Warde-Farley, S.~Ozair,
  A.~Courville, Y.~Bengio, Generative adversarial nets, in: Advances in neural
  information processing systems, 2014, pp. 2672--2680.

\bibitem{deng2014deep}
L.~Deng, D.~Yu, Deep learning: methods and applications, Foundations and trends
  in signal processing 7~(3--4) (2014) 197--387.

\bibitem{kim2017convolutional}
P.~Kim, Convolutional neural network, in: MATLAB deep learning, Springer, 2017,
  pp. 121--147.

\bibitem{rezagholizadeh2018semi}
M.~Rezagholizadeh, M.~A. Haidar, Semi-supervised regression with generative
  adversarial networks for end to end learning in autonomous driving (2018).

\bibitem{olmschenk2018generalizing}
G.~Olmschenk, Z.~Zhu, H.~Tang, Generalizing semi-supervised generative
  adversarial networks to regression, arXiv preprint arXiv:1811.11269 (2018).

\bibitem{aggarwal2019benchmarking}
K.~Aggarwal, M.~Kirchmeyer, P.~Yadav, S.~S. Keerthi, P.~Gallinari, Benchmarking
  regression methods: A comparison with cgan, arXiv preprint arXiv:1905.12868
  (2019).

\bibitem{nguyen2021fuzzy}
R.~Nguyen, S.~K. Singh, R.~Rai, Fuzzy generative adversarial networks (2021).
\newblock \href {http://arxiv.org/abs/2110.14588} {\path{arXiv:2110.14588}}.

\bibitem{mirza2014conditional}
M.~Mirza, S.~Osindero, Conditional generative adversarial nets, arXiv preprint
  arXiv:1411.1784 (2014).

\bibitem{van2020analyzing}
E.~van Krieken, E.~Acar, F.~van Harmelen, Analyzing differentiable fuzzy logic
  operators, arXiv preprint arXiv:2002.06100 (2020).

\bibitem{kingma2014adam}
D.~P. Kingma, J.~Ba, Adam: A method for stochastic optimization, arXiv preprint
  arXiv:1412.6980 (2014).

\bibitem{qiu2006wavelet}
H.~Qiu, J.~Lee, J.~Lin, G.~Yu, Wavelet filter-based weak signature detection
  method and its application on rolling element bearing prognostics, Journal of
  sound and vibration 289~(4-5) (2006) 1066--1090.

\bibitem{CHEN201916}
A.~Chen, T.~R. Kurfess,
  \href{https://www.sciencedirect.com/science/article/pii/S0888327018301250}{Signal
  processing techniques for rolling element bearing spall size estimation},
  Mechanical Systems and Signal Processing 117 (2019) 16--32.
\newblock \href {https://doi.org/https://doi.org/10.1016/j.ymssp.2018.03.006}
  {\path{doi:https://doi.org/10.1016/j.ymssp.2018.03.006}}.
\newline\urlprefix\url{https://www.sciencedirect.com/science/article/pii/S0888327018301250}

\bibitem{dragomiretskiy2013variational}
K.~Dragomiretskiy, D.~Zosso, Variational mode decomposition, IEEE transactions
  on signal processing 62~(3) (2013) 531--544.

\end{thebibliography}
\end{document}